\documentclass[sigconf]{acmart}
\AtBeginDocument{%
    \providecommand\BibTeX{{%
            \normalfont B\kern-0.5em{\scshape i\kern-0.25em b}\kern-0.8em\TeX}}}

\setcopyright{acmcopyright}
\copyrightyear{2022}
\acmYear{2022}
\acmDOI{XXXXXXX.XXXXXXX}

\usepackage{bbm}
\usepackage{caption}
\usepackage{subcaption}
\usepackage{amsmath}
\usepackage{color}
\usepackage{algorithm}
\usepackage{algpseudocode}
\usepackage{xspace}

\acmConference[WWW '23]{THE
    WEB CONFERENCE 2023}{April 30--May 04,
    2023}{Austin, TX, USA}
\acmPrice{15.00}
\acmISBN{978-1-4503-XXXX-X/18/06}

\begin{document}

\newcommand{\mt}{\ensuremath{\mu}-topic\xspace}
\newcommand{\mts}{\ensuremath{\mu}-topics\xspace}
\newcommand{\MTS}{\ensuremath{\mu}-Topics\xspace}

    \title{Producing Usable Taxonomies Cheaply and Rapidly at Pinterest Using Discovered Dynamic \MTS}
    %\subtitle{}

    \author{Abhijit Mahabal}
    \email{amahabal@gmail.com}
    \orcid{0000-0001-6911-2025}
    \author{Jiyun Luo}
    \email{jluo@pinterest.com}
    \affiliation{%
        \institution{Pinterest Inc.}
        \streetaddress{651 Brannan St}
        \city{San Francisco}
        \state{California}
        \country{USA}
        \postcode{94107}
    }

    \author{Rui Huang}
    \email{ruihuang@meta.com}
    \affiliation{
        \institution{Meta Inc.}
    }

    \author{Michael Ellsworth}
    \email{mellsworth@pinterest.com}
    \author{Rui Li}
    \email{rli@pinterest.com}
    \affiliation{%
        \institution{Pinterest Inc.}
        \streetaddress{651 Brannan St}
        \city{San Francisco}
        \state{California}
        \country{USA}
        \postcode{94107}
    }


    \begin{abstract}
Creating a taxonomy of interests is expensive and human-effort intensive: not only do we need to identify nodes and interconnect them, in order to use the taxonomy, we must also connect the nodes to relevant entities such as users, pins, and queries. Connecting to entities is challenging because of ambiguities inherent to language but also because individual interests are dynamic and evolve.

Here, we offer an alternative approach that begins with bottom-up discovery of \mts called pincepts. The discovery process itself connects these \mts dynamically with relevant queries, pins, and users at high precision, automatically adapting to shifting interests. Pincepts cover all areas of user interest and automatically adjust to the specificity of user interests and are thus suitable for the creation of various kinds of taxonomies. Human experts associate taxonomy nodes with \mts (on average, 3 \mts per node), and the \mts offer a high-level data layer that allows quick definition, immediate inspection, and easy modification. Even more powerfully, \mts allow easy exploration of nearby semantic space, enabling curators to spot and fill gaps.  Curators' domain knowledge is heavily leveraged and we thus don't need untrained mechanical Turks, allowing further cost reduction. These \mts thus offer a satisfactory ``symbolic'' stratum over which to define taxonomies. We have successfully applied this technique for very rapidly iterating on and launching the home decor and fashion styles taxonomy for style-based personalization, prominently featured at the top of Pinterest search results, at 94\% precision, improving search success rate by 34.8\% as well as boosting long clicks and pin saves.
    \end{abstract}

    \keywords{taxonomy, micro-topics, topic-mining, computer assisted data curation}

\begin{teaserfigure}
    \begin{subfigure}[b]{\textwidth}
        \centering
        \includegraphics[width=\textwidth]{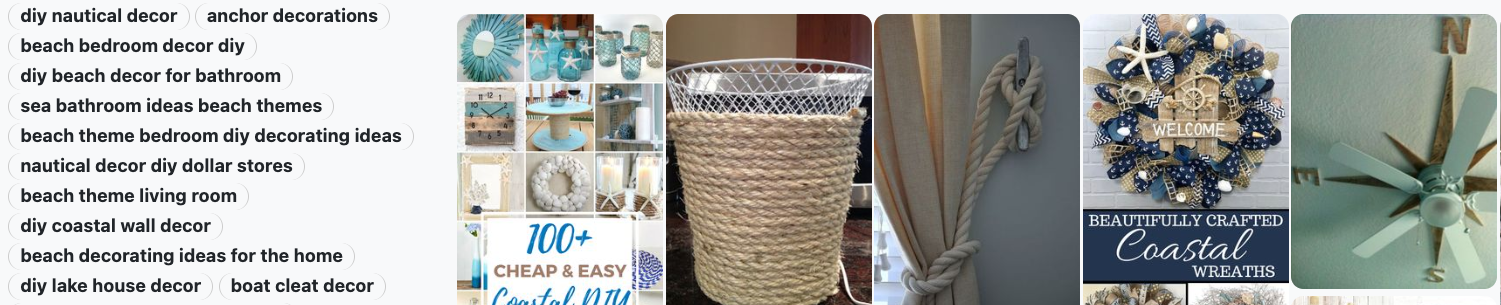}
        %\caption{Style recommendations for ``living room ideas''}
        %\label{fig:five over x}
    \end{subfigure}
    %\hfill
    \caption{An example of a discovered \mt{} displaying associated queries and pins. Notice the wide vocabulary of the queries, with terms like \emph{nautical, anchor, sea, beach, coastal, boat cleat} and so on. Taxonomy nodes are defined as a composition of on average three such \mts.}
    \label{fig:examples}
\end{teaserfigure}

    \maketitle

    \newcommand{\todo}[1]{{\color{red}[[\scshape Todo: #1]]}}
    \newcommand{\cmt}[1]{[{\color{blue}[\scshape Note: #1]}]}
    \newcommand{\query}[1]{\emph{[#1]}}

    \section{Introduction}

Human-curated taxonomies are important in the industry and power many applications such as web search \cite{Yin2010}, product recommendation \cite{Ziegler2004}, and personalization \cite{Zhang2014}, as well as exploration of a semantic space, such as here. However, creating and deploying taxonomies is expensive. The cost comes not just from the expertise needed in identifying nodes but even more from the data-hungry creation of machine-learning models needed to make the taxonomy usable.

To make a taxonomy usable, we must be able to connect it to relevant entities such as users and queries, and in the case of Pinterest, with the user-created images called ``pins''. The main example we use here to showcase our proposed technique is that of home decor and fashion styles. When a user issues a broad home decor query which can be satisfied in many different styles (such as the query ``bathroom organization'', whose results can span a wide range of styles), it is useful to show results in styles that a user cares about. To be able to do this task well, we must be able to identify styles with pins and users.

Put differently, taxonomy nodes, by themselves, are \textit{symbolic}. The node's name alone does not contain enough information to identify relevant entities. For example, the node ``nautical'' in our taxonomy, by itself, is of no use unless we can identify users who may be interested in it or when a pin or query refers to this style. This identification is a challenging problem, since the pin may use completely different terms such as \emph{coastal, beach theme, ocean decor} or \emph{marine decor}. The machine-learning for such connections, via data gathering and model training, is the most expensive part of making the taxonomy usable.

The problem is rendered more complex because of evolution of topics. The terminology associated with a topic changes over time as new words and entities enter language. A topic about a sports team, for instance, can enlarge when a new team member joins. Repeated data gathering to keep track of changing topics adds to costs and is rarely done in practice.

We offer an alternate process. By a bottom-up statistical analysis of query sessions, we generate a symbolic layer of \mts called Pincepts, where each \mt already comes equipped with thousands of queries from a rich vocabulary, as well as thousands of related pins. The role of the human experts is to attach \mts for each taxonomy node: on average, we needed about three \mts per taxonomy node. Furthermore, the \mts are dynamic: the associated pins and queries automatically adjust to the changing world. The effort is thus reduced from finding and labeling potentially hundreds of entities per node to just two or three easy-to-find \mts.

Of course, this whole endeavor crucially depends on the expressive power of the discovered \mts. For the taxonomies we built, we found the \mts to be adequately expressive. The simple fact is that any meaningful real category will surely have some sessions dedicated to it, out of the billions of sessions we see. Our algorithms adequately deal with arbitrary mixtures of the wildly popular and the quaintly obscure.

The generated \mts are independent of the taxonomy and can thus support many different taxonomies. Indeed, apart from the styles taxonomy example seen here, we are rebuilding our multi-thousand-node taxonomy on exactly the same set of \mts, with automated algorithmic mappings between taxonomy nodes and \mts.

Our process to reducing efforts contrasts with another research program also aimed at reducing effort: automated taxonomy generation \cite{Zhang2018a,Shen2022,Lee2022a}. There, the taxonomy is automatically generated and important dimensions automatically selected. Here, we retain human agency in creating topics, thereby controlling what aspects we care about, such as, here, styles, preferring to reduce human effort by making that process smoother. An automatic taxonomy generator is perhaps unlikely to consider style to be the salient dimension and thus cannot lead to the taxonomy we seek.
    \section{Generating \MTS}

\newcommand{\pic}[1]{\raisebox{-.4in}{\includegraphics[height=0.5in]{pic/#1.png}}}

\begin{table*}[t]
    \caption{Illustrative Examples of Pincepts. The top queries are displayed for each. Notice the wide-ranging vocabulary of dog names in example 1; the nuanced distinctions among intents in examples 2a--2c each of which involves \emph{garlic} but differently; and, in examples 3a and 3b, broad and much narrower topics from Old Hollywood are seen.}
    \centering
    \label{tab:pincepts}
    \begin{tabular}{cp{5.53in}p{1in}}
        1&\query{australian shepherd}, \query{border collie}, \query{german shepherd}, \query{golden retriever}, \query{australian shepherd puppy}, \query{dog breeds}, \query{german shepherd puppies}, \query{husky}, \query{golden retriever puppy}, \query{bernese mountain dog}, \query{labrador retriever}, \query{puppies}, \query{husky puppies}, \query{collie}, \query{border collie puppies}, \query{collie dog}, ...&\pic{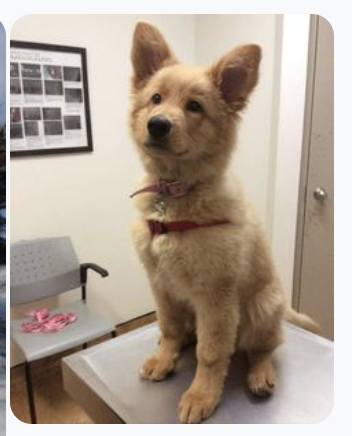}\\
        2a&\query{how to grow garlic}, \query{grow garlic indoors}, \query{grow garlic from clove}, \query{plant garlic from clove}, \query{growing garlic}, \query{planting garlic}, \query{regrow garlic}, \query{growing garlic from cloves in water}, \query{grow garlic from clove in pots}, ...&\pic{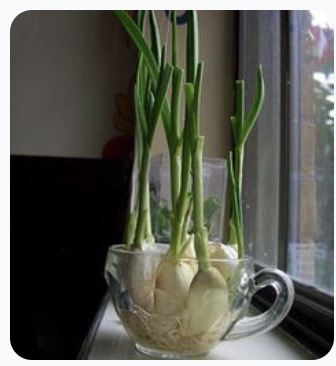}\\
        2b&\query{how to store garlic}, \query{how to store garlic cloves}, \query{how to store garlic bulbs}, \query{best way to store garlic}, \query{garlic storage}, \query{store garlic}, \query{storing garlic cloves}, \query{freezing garlic}, \query{how to keep garlic fresh how to store}, \query{store garlic cloves how to}, \query{how to preserve garlic},  \query{how to keep garlic fresh}...&\pic{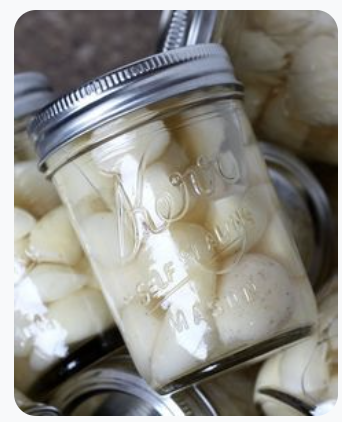}\\
        2c&\query{garlic knots}, \query{easy garlic knots}, \query{homemade garlic knots}, \query{garlic knots recipe}, \query{garlic knots recipe easy}, \query{garlic knots pizza dough}, \query{garlic knots recipe from scratch}, \query{garlic knots crescent rolls}, \query{garlic knots pillsbury}, \query{garlic bread}, \query{garlic rolls}, \query{quick garlic knots}, \query{homemade garlic knots easy}, ...&\pic{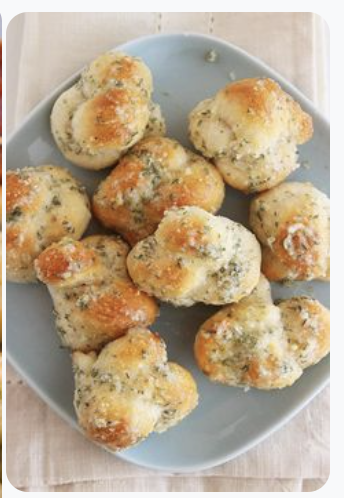}\\
        3a&\query{cary grant}, \query{gregory peck}, \query{gary cooper}, \query{rock hudson}, \query{montgomery clift}, \query{paul newman}, \query{clark gable}, \query{cary grant young}, \query{marlon brando}, \query{james stewart}, \query{errol flynn}, \query{old hollywood actors}, \query{tyrone power}, \query{jimmy stewart}, \query{classic hollywood men}, \query{humphrey bogart}, \query{james dean}, ...&\pic{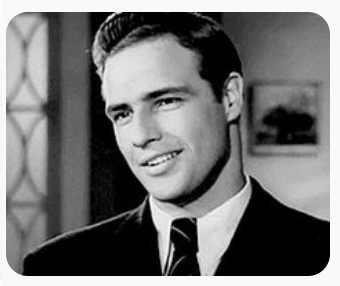}\\
        3b&\query{marilyn monroe drawing}, \query{marilyn monroe painting}, \query{marilyn monroe pop art}, \query{marilyn monroe artwork}, \query{marilyn monroe art}, \query{marylin monroe art}, \query{marylin monroe}, \query{marilyn monroe stencil}, \query{marilyn monroe tattoo}, \query{marilyn monroe sketch}, \query{marilyn monroe wallpaper}, ...&\pic{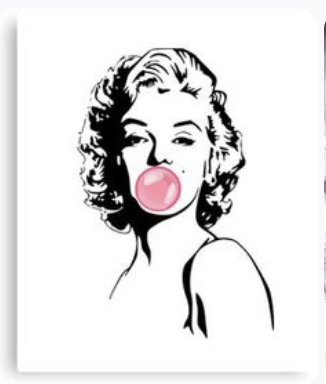}\\
    \end{tabular}
\end{table*}

Our \mts are generated with zero-supervision using only search query sessions, where each query session is a set of queries issued by a single user in a short span of time. Query sessions thus tend to, statistically speaking, stick to a single topic. Our algorithm discovers as a \mt any topic that is session-coherent: that is, topics to which entire sessions are dedicated. It so happens that most meaningful topics, including individual styles we aimed for to enable style-based personalization, are session-coherent since some users will make a series of queries on that topic.

It is important to note that the clustering-based algorithm shown here is one specific algorithm that gets the job done. We chose this algorithm non-arbitrarily, but we recognize that other algorithms and other clustering techniques may also work well.

\newcommand{\graph}{\ensuremath{\mathcal{G}}}
\newcommand{\bipartite}{\ensuremath{\mathcal{B}}}

The method is described here briefly given the limited space. We begin with a bipartite graph \bipartite{} of queries and n-grams generated from sessions. This is exactly the graph used in \cite{Mahabal2020}, where the construction is fully spelled out. Each n-gram can now be seen as a distribution over queries: for an n-gram $n$, it is associated with a query $q$ if that query is seen in sessions that mention the $n$. Notice that this definition implies that the query distribution of $n$ is dynamic: if we recreate the bigraph over a different time window, a different set of queries will be picked up.

From \bipartite{} we create a weighted graph \graph{} whose nodes are the n-grams, and we create an edge between two n-grams with weight based on the continuous Jaccard similarity of the associated query distributions. Only edges with Jaccard similarity over a threshold (0.3; somewhat arbitrarily chosen by looking at data) are retained.

If the edge-weight between an n-gram $n_i$ and query $q_j$ is represented by $W_{n_i,q_j}$, and the set of all queries is $Q$, then the continuous Jaccard similarity between $n_i$ and $n_k$ is simply

\begin{equation}
    \frac{\sum_{q\in Q}\min({W_{n_i, q}, W_{n_k, q}})}{\sum_{q\in Q}\max({W_{n_i, q}, W_{n_k, q}})}
\end{equation}

As is obvious, this is a generalization of the traditional Jaccard similarity where each $W_{n, q}$ is 0 or 1.

On graph \graph{} we do community discovery. Our communities are semi-cliques, that is, densely connected subgraphs, building on \cite{Madani2010}. The notion of creating a catalog of topics based on clustering is also explored by \cite{Panchenko2017}. In particular, \cite{Panchenko2017} emphasizes the role of similar words as a way of teasing apart senses and doing automated disambiguation. We use the Chinese Whispers Algorithm \cite{Biemann2006} for cluster discovery in ego-neighborhoods of each node of a graph, but with a simulated-annealing tweak, where we start with only the highest weighted edges in the early iterations and slowly introduce weaker edges in subsequent iterations. Communities discovered from each ego-neighborhood are combined and de-duplicated to produce our \mts. Other ways of community discovery should also work provided that they, like here, produce overlapping clusters.

Each \mt, thus, is a set of n-grams. It is also, at any time, an ordered list of queries obtained from \bipartite{} by looking at queries most strongly associated with these n-grams, or pins whose description contains these n-grams, and users who interacted with these pins. The human experts use the associated queries and pins to determine if a \mt is a good fit for a taxonomy node, but what they are really choosing, unbeknownst to them, is the set of n-grams. When \bipartite{} is recalculated at a different time, the associated pins and queries will have dynamically updated.

Examples of produced clusters are shown in Table 1, and the caption there underscores the semantic properties of the generated \mts.

\subsection{Range and Limitations of \MTS}

\MTS are discovered from tokens which come from sessions. Anything that is frequently the subject of sessions will get discovered as a \mt, and this includes all interests. There are topics one could ``define'' that are not really searched for together and will thus not show up as a pincept. An example is the made up concept from Alice in Wonderland ``things that begin with M (such as mousetraps, the moon, memory, and muchness)'': this will not show up as a pincept since these things are not ``session cohesive''.

More practical examples of topics that are not session-cohesive include obscure specific entities that people search not in isolation but as part of a wider topic. One example is the Hollywood actor James Dean (1931--1955): although people search for him, it is usually as part of a wider topic, such as Hollywood of the 1950s, or the even wider ``old Hollywood'' and these wider \mts are picked up, and they include James Dean but also Marlon Brando, Warren Beatty, Rita Hayworth, Marilyn Monroe and so forth. By contrast, people certainly search specifically for Marilyn Monroe and for Marlon Brando (i.e, they are the topic of search in their own right rather than just an exemplar of something broader), and the algorithm picks up dedicated pincepts for these.

We think this disparity in granularity is a feature, not a bug: our \mts auto-adjust in their granularity to what topics users search for. If different users focus on very different nuances of a single topic (such as garlic and its nuances, examples 2a--2c in the table), the algorithm captures these as distinct topics, whereas if the distinction among members of some category seems unimportant to our users, it can result in a wider, undifferentiated topic (such as example 3a in the table).

    \section{Deployment and Impact}

We used the aforementioned techniques to produce taxonomies for styles, one for Home Decor and one for Fashion. The taxonomy nodes here are complex with huge vocabularies that overlap (such as between the HD styles \emph{French Country} and \emph{Parisian Modern}).

\subsection{Taxonomies and Associated Micro-topics}
The number of nodes in the taxonomies is around one hundred each: this was enough to capture the major styles we wished to capture. It is important to highlight how tiny the manual work involved here is: each taxonomy node was defined as a set of \mts, and this could be done rapidly given how few associations needed to be made. For fashion, the number of associated \mts had a mean of 3.36 and a mode of 2, while for home decor these numbers were 2.88 and 2. Less than 2\% cases required more than 6 \mts. We attached the most \mts, 13, with the highly multifarious ``kids theme'' in home decor: although we selected broad \mts covering this whole style, we also found specific \mts covering in depth some sliver of this style, including variations for boy's room vs girl's room as well as rooms shared by siblings, as well as pincepts about specific themes such as unicorn, rainbow, superhero and pirates.

\subsection{The Curatorial Process}

Four domain experts (two each for fashion and HD) decided on the styles they wanted. In just two days, the experts mapped the style nodes to the \mts. The data-assisted tools helped them discover additional styles to cover as well as enrich the nodes they had, as described below.

To illustrate the process, imagine that the domain expert is considering the HD style ``nautical''. In our tool, they look up a relevant seeming query and its associated \mts. The query \query{nautical decor} is associated with 12 \mts. A fraction of the tool's user interface is shown in Figure \ref{fig:examples}, where for each \mt, the expert can see the top queries and the top pins, making it very easy to decide if the \mt is relevant to the taxonomy node under consideration. It is important to not choose a \mt broader than the taxonomy node, although it is perfectly fine to choose a narrower \mt.

One \mt the experts chose for \textit{nautical} has a range of queries showcasing the many ways a nautical style can be evoked even without using the term \textit{nautical}. The queries make use of terms such as \textit{beach decor}, \textit{coastal home}, \textit{beach theme}, \textit{anchor theme}, \textit{ocean decor}, \textit{ship decor}, \textit{shark theme}, \textit{lake house decor}, and \textit{marine decor}, to name just a few.

\begin{figure}[t]
    \centering
    \includegraphics[width=0.47\textwidth]{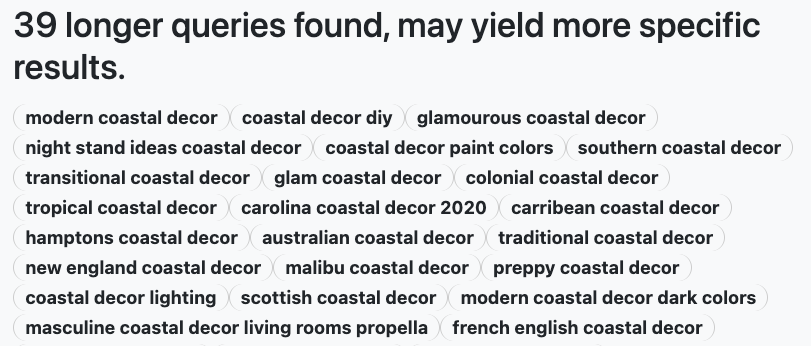}
    \caption{For the query \query{coastal decor}, we see additional queries displayed in the UI. These queries have associated \mts beyond those of the original query and may offer possibility of better covering a specialized area of the relevant semantic domain.}
    \label{fig:additional}
\end{figure}

Apart from seeing and selecting displayed \mts, the expert may also use more specific queries offered by the UI (and shown in \ref{fig:additional}). These are queries with additional words that we know to be linked to additional \mts, such as a specialized area of the semantic space we are trying to cover. For \query{coastal decor}, 39 more specialized queries are shown, and the human expert can look over them to see if they can provide wider coverage. These include Australian, Caribbean, and Hampton specific regional variants, interactions with other styles (such as with Glam and Modern), and so forth.

Some \mts for a query may not be relevant to the taxonomy node because they are too broad or focus on aspects not directly related to style. For instance, one \mt was about external views of beach houses, but the taxonomy is about interior design. These \mts can simply be ignored and not attached to any taxonomy node: the human-in-the-loop paradigm keeps the mappings clean. Another \mt we skipped for \textit{nautical} was pirate-themed: it was instead assigned to the \textit{Kids' Themed} style.

Using these methods, \mts are manually hooked to the taxonomy node. Since each \mt represents many hundreds of queries and many hundreds of pins, in a matter of minutes, we have associated thousands of queries and pins with the taxonomy node, allowing very robust downstream processes as described below. Moreover, the process may reveal other taxonomy nodes that are needed, those that were not originally considered by the curators, making the taxonomy itself more robust.

Importantly, when queries are shown in the UI, we also display how popular these queries are. This allows the curator to make a decision informed by actual user demand rather than an imagined user demand.

\subsection{Triggering}

Since a taxonomy node is a set of \mts and we know the pins and queries associated with each \mt, we can also calculate the styles a user may be interested in based on the pins they have interacted with or the queries they have issued. For example, some user may have saved pins or searched for farmhouse home decor style, and they have thus implicitly expressed an interest in this style. Now, when they search for a generic home decor query (say, \query{kitchen ideas}), it is possible that they would like images that match their query and are also in the styles they care about.

For broad queries, we are able to find user-relevant styles and many pins in the intersection of ``relevant to the query'' and ``consistent with the style''. We enable navigation by style: at the top of Pinterest search results, we prominently display ``styles for you'', where the styles are specific to the user and the pins displayed within each style are specific to both the query and the style.

\subsection{Offline Quality Measurement}

Our taxonomies have two levels: a top level of style (such as \textit{French} and \textit{Industrial}) and then sub-styles (\textit{French Country} and \textit{Parisian Modern} for the former; \textit{Loft} and \textit{Steampunk} for the latter).

This is a challenging domain to classify because not only do styles make use of similar words (all of them talk about bedrooms and bathrooms, for instance), there are plenty of ``leakages'' and fusions among styles. Thus, people talk of ``nautical farmhouse decor'' but also \textit{Japandi}, a Japanese and Scandinavian inspired style. This is even more true in fashion styles, where experimentation is much cheaper and faster than in home decor. Contrast the difficulty here of distinguishing the style \textit{French Country} from the style \textit{Parisian Modern} with the relative simplicity of telling apart a politics news story from a sports news story \cite{Meng2020}.

We measured our quality by randomly sampling 50 pins for evaluation from the top 1000 highest scoring pins for each style. Home Decor and Fashion specialists checked the predicted style and sub-style. For both fashion and for Home Decor, the classifier achieved high precision for style (91.4\% and 90.4\%, respectively). For sub-style, the performance was less stellar: 61\% and 78\% respectively. These two sets of figures imply that when the classifier made an error on a sub-style, it still stayed within the broader style.

\subsection{Live Quality Measurement}

On live traffic, our human evaluation shows that we suggest applicable styles for 94\% of the time. Our module triggers for around 16\% of all Fashion and Home Decor queries, and the open rate (here 9.5\%) and search success rate (a measure of how long they stay and explore; here 31\%, which is 34.8\% higher than the prior success rate of the same queries) is the highest for search modules at Pinterest; the next best module has 6.1\% and 21.3\%, respectively.

    \bibliographystyle{ACM-Reference-Format}
    %\balance
    \bibliography{bibs}

\end{document}